\documentclass{article}

\usepackage{arxiv}
\usepackage{listings}
\usepackage{framed}
\usepackage{xcolor}  
\usepackage{calc}
\usepackage{float}  
\usepackage[utf8]{inputenc} 
\usepackage[T1]{fontenc}    
\usepackage{hyperref}       
\usepackage{url}            
\usepackage{booktabs}       
\usepackage{amsfonts}       
\usepackage{nicefrac}       
\usepackage{microtype}      
\usepackage{lipsum}
\usepackage{graphicx}
\graphicspath{ {./images/} }

\title{%
Imitating Mistakes in a Learning Companion AI Agent for Online Peer Learning%
\thanks{This work has been published in the Proceedings of the 2025 19th International Conference on Ubiquitous Information Management and Communication (IMCOM). DOI: \href{https://doi.org/10.1109/IMCOM64595.2025.10857528}{10.1109/IMCOM64595.2025.10857528}. © 2025 IEEE.}%
}

\author{
 Sosui Moribe \\
  Gratuate School of Design\\
  Kyushu University\\
  Fukuoka, Japan \\
  \texttt{moribe.sosui.695@s.kyushu-u.ac.jp} \\
   \And
 Taketoshi Ushiama \\
  Faculty of Design\\
  Kyushu University\\
  Fukuoka, Japan \\
  \texttt{ushiama@design.kyushu-u.ac.jp} \\
}

\begin{document}

\maketitle

\begin{abstract}
In recent years, peer learning has gained attention as a method that promotes spontaneous thinking among learners, and its effectiveness has been confirmed by numerous studies. This study aims to develop an AI Agent as a learning companion that enables peer learning anytime and anywhere. However, peer learning between humans has various limitations, and it is not always effective. Effective peer learning requires companions at the same proficiency levels. In this study, we assume that a learner's peers with the same proficiency level as the learner make the same mistakes as the learner does and focus on English composition as a specific example to validate this approach.
\end{abstract}


\section{Introduction}
In recent years, active learning has gained importance in education as it fosters spontaneous thinking and understanding through self-directed activities\cite{freeman2014active}. Peer learning, a form of active learning, involves learners cooperating with peers. Combining "Peer" and "Learn," it enables learners to teach and receive feedback from each other rather than relying solely on teachers. This approach deepens understanding while enhancing collaborative and communication skills. Due to these benefits, peer learning is widely adopted across various subjects.

On the other hand, online learning environments have enabled learners to study anytime and anywhere\footnote[1]{Khan Academy, \url{https://en.khanacademy.org/}}. 
In particular, a large number of tutoring services, one-shot online courses, and open course platforms offered by universities and companies provide diverse learning opportunities.
Similarly, e-learning platforms that allow students to solve numerous exercises play an important role in education. In most cases, learners learn by watching video lectures online and solving problems provided by the website. 

\begin{figure}[tbp]
\centering
\includegraphics[width=\columnwidth]{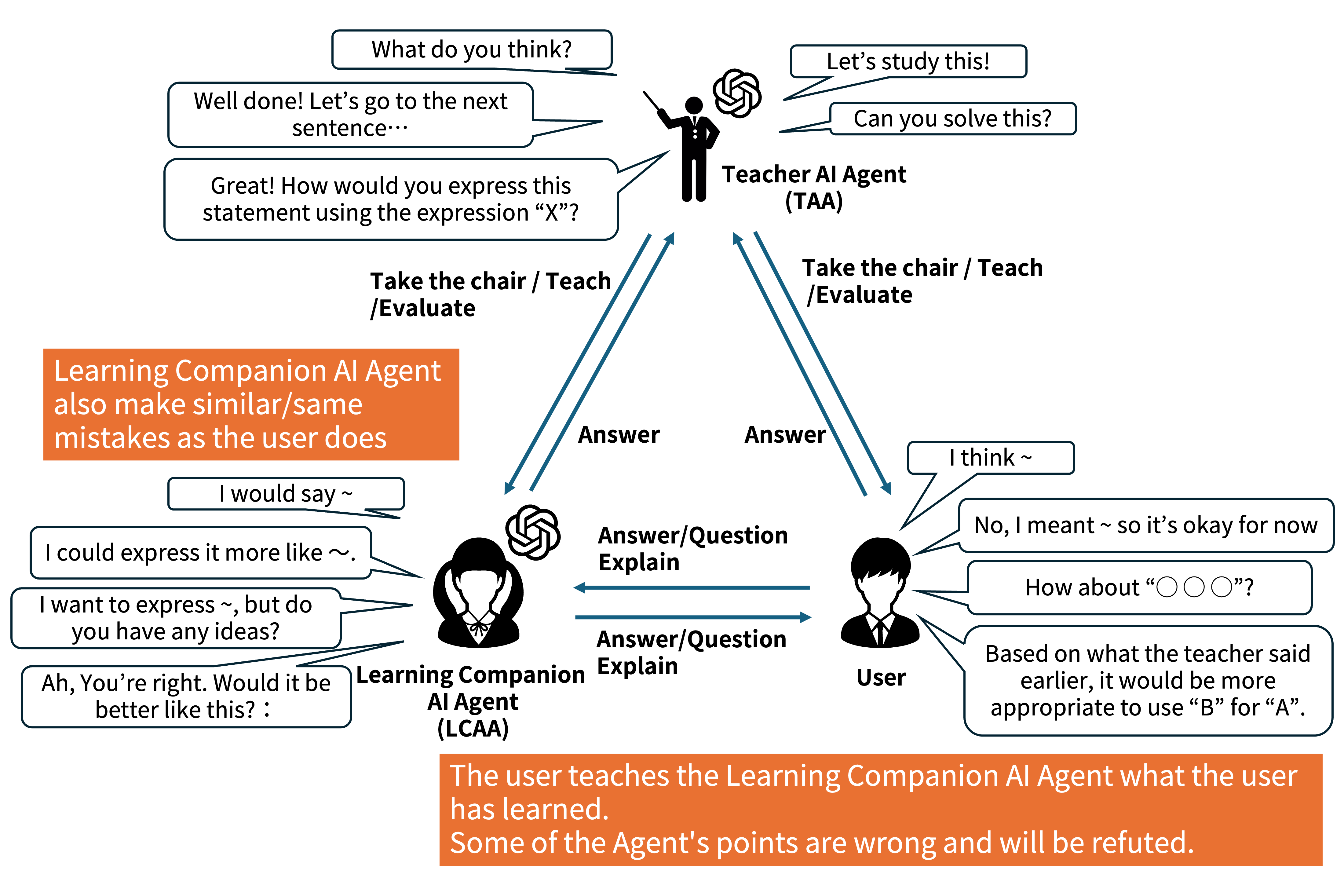}
\caption{Agent-Based Online Peer Learning Environment}
\label{figSys}
\end{figure}

Another important aspect of the modern learning environment is the use of Large Language Models (LLMs), such as GPT-4 and Llama3, which are now widely used across various fields. 
For example, GPT-4 has demonstrated advanced reasoning and language abilities by passing the U.S. bar exam \cite{openai2024gpt4}. These LLMs can be utilized as interactive applications, education is one of the key areas their use is expanding. A common approach is to use LLMs take on teaching roles in learning environments. Interactive agents, referred to as “AI Agents”, powered by LLMs are now widely adopted, enabling learners to engage in education at low cost.

We are developing a learning environment with AI Agent for online peer learning.
Fig. \ref{figSys} provides an overview of this environment. It consists of a Teacher AI Agent (TAA), a Learning Companion AI Agent (LCAA), and a User.
The TAA facilitates the user and the LCAA by giving assignments and asking the other learner's opinion about one learner's answers.
The User interacts with the LCAA by asking for feedback and identifying mistakes in each other's responses.

LCAA is crucial in this environment. 
Since the learner needs to correct the LCAA's mistakes, which mirror errors the learner has overcome, it is important that the LCAA generate errors that the learner would make.
Therefore, to realize effective peer learning, it can be considered that the LCAA must have the same level of proficiency as the users. 
In peer learning, learners explain the reasoning behind their knowledge and ideas when teaching other learners. This process not only strengthens their understanding but also helps them identify gaps in their own knowledge. Through discussions and responses to peer questions or counterarguments, learners generate new insights that improve learning outcomes \cite{tullis2020peer}.
If AI Agents can be realized as learning companions, they can facilitate effective peer learning anytime, anywhere in an online environment, completing the existing role of AI Agents as teachers. 

In this study, we propose an AI-based learning environment that combines the strengths of both approaches, addressing their limitations by using AI Agents as learning companions. As a concrete example, we focus on English composition. The contributions of this study are$\colon$

\begin{itemize}
\item proposing an AI-based Peer Learning environment,
\item proposing a method for the LCAA to generates errors based on the learner's proficiency level, and
\item demonstrating the effectiveness of this approach in generating user-like mistakes in English composition.
\end{itemize}

The remainder of this paper is organized as follows.
In Section II, we review related works on peer learning and AI in educational contexts.
Section III outlines the proposed peer learning environment and describes the functions and roles of the Learning Companion AI Agent.
Section IV details the experimental design and method, followed by the results and analysis in Section V.
Finally, in Section VI, we present our conclusions and discuss potential directions for future work.

\section{Related Works}
Various methods have been explored for developing AI Agents for learning companions. In this section, we review key research on tuning AI Agents and practical applications of AI in learning environments.

\subsection{Peer Learning}

There are various types of peer learning, which have different names depending on the implementation methods and the subject matter being addressed, and in particular, \cite{abdelkarim2016peer} and \cite{gubalani2023peer} have shown the effectiveness of Peer Tutoring in the areas of mathematics and English reading, respectively.

A learning method where multiple learners correct each other's compositions and discuss them is called Peer Review. While there are variations such as Peer Response, Collaborative Translation depending on the subject matter and the method, in this study, we use the term 'Peer Review' to cover these variations. 

In English as a Foreign Language (EFL) environments, such as Japan, many efforts have been made to implement Peer Review of English compositions in school classes \cite{adlan2020collaborative}\cite{ono2019peer}\cite{suzuki2022practical}\cite{yoshizawa2010peer}\cite{berg1999trained}.  Similarly, in other languages, many studies have demonstrated the effectiveness of Peer Review for learning non-native languages\cite{mochizuki2013peer}\cite{sanchez2019peer}.
Not only is the learning effect significant, but also that 93\% of learners also express a desire to receive peer feedback \cite{jacobs1998feedback}. Some studies have shown that peer learning improves the motivation\cite{mochizuki2013peer}\cite{sanchez2019peer}, and highlighting its usefulness from multiple perspectives.

\subsection{Online Learning}
Online prep schools, where students prepare for high school and university entrance exams, are a common types of online learning environment\footnote{Toshin high school, \url{https://www.toshin.com/hs/}}.
These schools provide recorded lectures by instructors primarily covering subjects like English and mathematics for exam preparation.

There are also services that offer practical skills and one-off courses outside traditional exam subjects in a buy-out format\footnote{Coloso, \url{https://coloso.global/en}}\footnote{Udemy, \url{https://www.udemy.com/}}. These services enable private experts with experience in specific industries or creative skills such as computer graphics, drawing, design, programming, or cooking to offer courses on an individual basis, allowing students to learn a wide range of subjects according to their interests.

Online courses run by universities and companies are examples of courses that are run by organizations rather than individuals\footnote{Coursera, \url{https://www.coursera.org/}}\footnote{edX, 
\url{https://www.edx.org/}}, providing a wide variety of services. These programs offer courses and practice assignments in specialized fields accessible globally. Similarly, e-learning platforms that allow students to solve numerous exercises play an important role in learning\footnote{Free Code Camp, \url{https://www.freecodecamp.org/}}\footnote{Leet Code, \url{https://leetcode.com/}}. 
Due to the nature of the sites, each site is often specialized in the field it covers, but users may be able to reach a practical level through exercises, pass certification exams, or receive job offers from the certificates in the profiles that accompany the site.

\subsection{Challenges in Peer Learning and Online Learning}
Even if above learning environment and methods work well under ideal conditions, they may not always be effective for individual learning in practice.

For example, online learning faces the issue of high dropout rates. Experimental results \cite{ishii2017mooc} show that the completion rate of online video courses is only about 7.5\%, depending on the learner's background and the device used.
This suggests that unidirectional online learning may not effectively support continuous learning. 

Although many studies demonstrate the effectiveness of peer learning, it also presents several challenges. 
Peer learning environments are not always easily accessible, and learners experience psychological barriers, such as embarrassment about making mistakes in front of others, especially due to the need for direct human interaction \cite{fryer2006bots}. Thus, peer learning faces time and space constraints, as well as challenges in managing appropriate interactions between learners.

Reviewers may overlook mistakes, and both parties may not learn effectively from each other due to the lack of focus and tailored feedback when teaching others.
In other words, several studies have highlighted that peer learning itself requires a certain level of proficiency\cite{ono2019peer}\cite{yoshizawa2010peer}\cite{berg1999trained}\cite{mochizuki2013peer}\cite{sanchez2019peer}\cite{jacobs1998feedback}.

When learners' proficiency levels differ significantly, especially they lack metacognitive awareness of their understanding, it becomes challenging to engage in effective teaching and learning \cite{zou2023peer}.
For instance, \cite{berg1999trained}, in study examining the effects of Peer Review on learners with different language proficiency, participants are divided into upper and lower groups according to their TOEFL scores, and Peer Review was conducted with in each group. The study demonstrated the effectiveness of Peer Review in both groups. However, it also suggested that the effectiveness of Peer Review relies on the assumption that learning pairs have a similar level of proficiency levels.

To address these challenges, Chatbots (i.e., AI Agents) can provide continuous access to a learning environment, and reduce the embarrassment of making mistakes in front of others \cite{fryer2006bots}. Additionally, the common issue of learners relying too heavily on AI for thinking can be mitigated by incorporating peer learning, which encourages independent thinking and fosters a more effective learning environment.

However, since AI can easily provide plausible answers, there is a concern that learners may take the output for granted or become overly reliant on it \cite{kasneci2023chatgpt}\cite{lim2023generative}.
To address this, the Japanese Ministry of Education, Culture, Sports, Science and Technology has issued guidelines for using AI in education, recommending that teachers use to ensure that learners still engage in spontaneous in thinking \footnote{Tentative guidelines for the use of generative AI in the primary and secondary education stage, 2023, \url{https://www.mext.go.jp/content/20230718-mtx_syoto02-000031167_011.pdf}}.

\subsection{Modifying AI}
ChatEval\cite{chan2023chateval} shows that a multi-agent system can accurately evaluate sentences by enabling multiple agents, each assigned a distinct role in evaluating input text, to interact and debate.

Another significant study, Generative Agents\cite{park2023generative} seeks to replicate human behavior by assigning personas to AI agents. These independent AI agents are given personalities via prompts and act in a virtual space, where their actions are logged as memories, leading to spontaneous behavior. Optimizing of these agents based on logged actions are considered to have promising application in learning environments.

Chen et al. \cite{chan2024scaling} generate a large amount of synthetic persona data, called Persona Hub, based on the diverse persona profiles produces output with characteristics of these personas.
The study demonstrates the potential for creating various types of questions in mathematics, even if they require essentially the same calculations, and is considered one of the most promising methods for improving learning environments.

Chain-of-Thought prompting\cite{wei2022chain} is a method that improves performance on complex tasks by breaking down prompts into multiple steps, allowing the model to generate intermediate inference steps. This method is expected to engage output with the desired features for tasks that would be difficult to achieve with a single prompt for AI.

\subsection{Online Learning with AI}
A typical example of AI used in education before the advent of interactive AI Agent is an automatic essay grading system. Well-known systems include are Criterion\footnote{Criterion, \url{https://criterion.ets.org/default.aspx}} from ETS, which administers the TOEIC\footnote{TOEIC, \url{https://www.iibc-global.org/toeic/test/lr/guide01.html}} test, and Grammarly\footnote{Grammarly, \url{https://www.grammarly.com/}}, both of which provide automated writing corrections. Studies analyzing the use of automatic evaluation systems in EFL environment, we present \cite{saito2017automated} using Write \& Improve\footnote{Write \& Improve, \url{https://writeandimprove.com/}} and \cite{oda2017pigai} using Pigai\footnote{Pigai, \url{https://www.pigai.org/}}. Both studies show that the use of these systems can be positively impact learning by promoting spontaneous writing improvement. 
According to \cite{oda2017pigai}, automatic evaluation systems generally assess the similarity between the user's input and sentences written by native speakers based on a large corpus of data. These systems evaluate various aspects, including grammar, vocabulary usage, and sentence structure, to provide feedback.
Even when using AI Agents for teaching, the technology used in the automatic grading system is considered important because it is necessary to correctly evaluate the correctness of the learner's input and the way in which the learner makes mistakes.

\subsection{Learning with AI agent}
\subsubsection{Passive Learning}
In passive learning, it is common for learners to simply enter a question from a textbook or assignment and receive an answer, such as with ChatGPT\footnote{ChatGPT, \url{https://chatgpt.com/}}. Similarly, learners can organize their own thoughts and have their sentences proofread by AI Agents, and we believe that effective learning can occur if those tools are used properly.
However, as mentioned in the Introduction, some studies have raised concerns that the ease of the generative AI might lead learners to overly rely on it for answers.
\subsubsection{Active Learning}
SimStudent\cite{matsuda2013simstudent} is a system that facilitates learning by having students teach an AI Agent. In this system, an AI Agent called SimStudent attempts to solve a mathematical equation step by step. If the solution is incorrect, the learner points out the error and provides the correct method or formula. SimStudent learns from these interactions. If the learner misunderstands the solution, SimStudent will fail to reach the correct answer, highlighting gaps in the learner's understanding. This process ensure that the learner's grasp of the solution becomes more accurate.

\section{Proposed peer learning environment}

\subsection{Overview of the proposed learning envrionment}
\begin{figure}[tbp]
\centering
\includegraphics[width=0.9\linewidth]{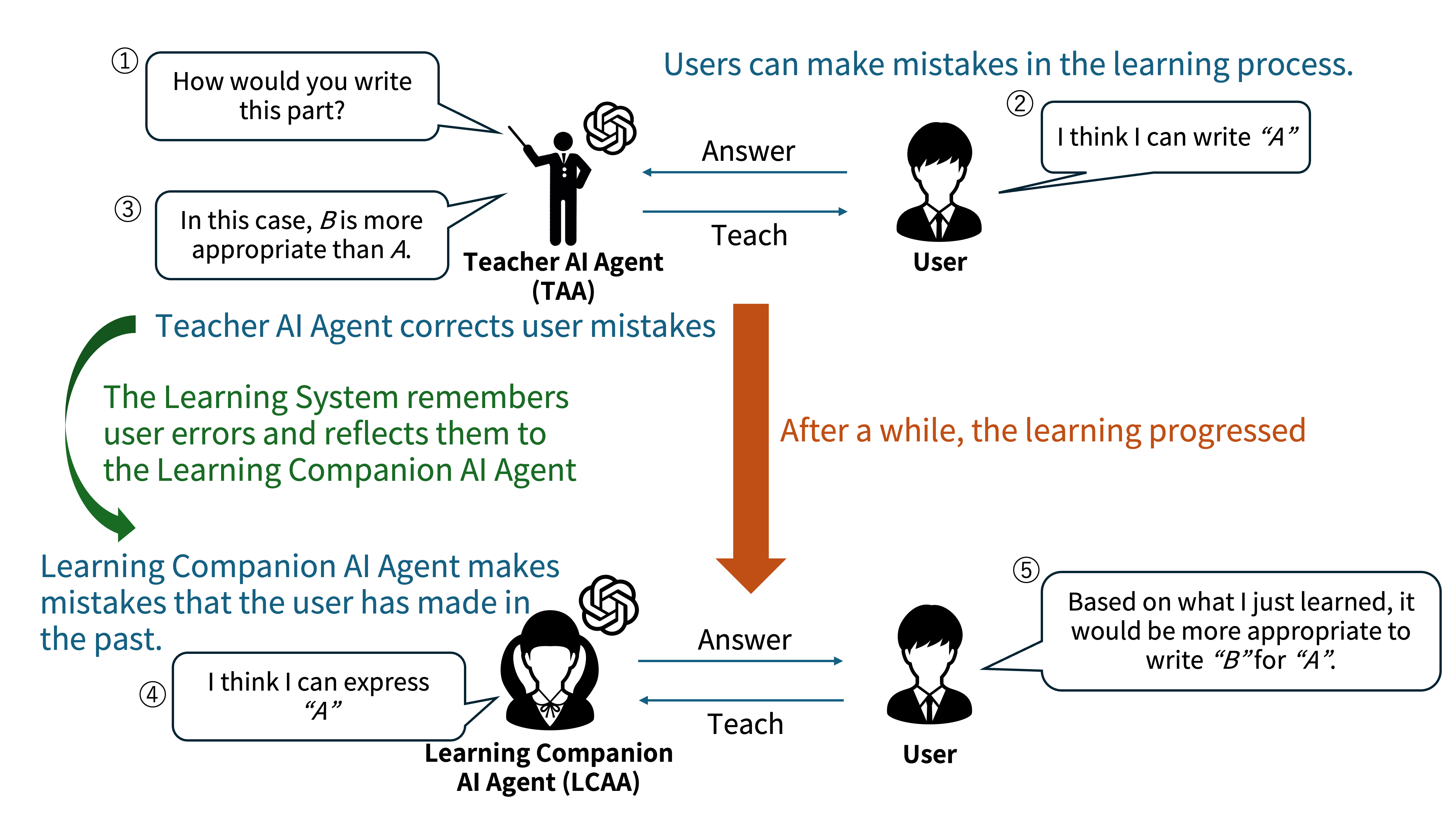}
\caption{Concept to realize a Learning Companion AI Agent at the same level as the user}
\label{figMiss}
\end{figure}

As discussed in Section II, although various peer learning methods exist, we propose a fundamental learning process within the framework of the proposed online peer learning environment. The learning flow is illustrated in Fig. \ref{figMiss}.

The typical learning flow envisioned in this study is outlined as follows$:$
\begin{enumerate}
\item The TAA submits questions to the user and LCAA.
\item The user answers the question.
\item The TAA corrects the user's mistakes and teaches or gives hint to solve the question.
\item The LCAA also answers the question.
\item The TAA instructs the user to correct the LCAA's mistakes based on the instruction TAA gave previously.
\item The user corrects the LCAA and discusses them.
\item The TAA concludes the discussion.
\end{enumerate}

\subsection{Functions required in LCAA}
The LCAA plays several roles in this peer learning environment. As discussed in Section II, for peer learning to be effective, learners generally need to have a similar proficiency level. 
Therefore, the following characteristics are essential for the LCAA$:$
\begin{itemize}
    \item Engages in conversations with users at an equal level, matching tone, experience, and knowledge.
    \item Detects errors at a similar proficiency level similar to the user.
    \item Produces writing of the same quality as the user's.
\end{itemize}

In this paper, we focus primarily on the LCAA’s ability to write at the same proficiency level as the user.
We assume that learners at the same proficiency level tend to make similar types of mistakes. 
For some learners, identifying and correcting mistakes that are too simple does not promote effective learning. 
Conversely, if the mistakes are too advanced, the learner may not notice them or may struggle to correct them. 
We hypothesize that learners consolidate their understanding most effectively by correcting mistakes that match their own level of difficulty and by identifying and addressing similar mistakes made by others.

\subsection{Generating an answer including mistakes}

\begin{figure}[tbp]
\centering
\includegraphics[width=0.6\linewidth]{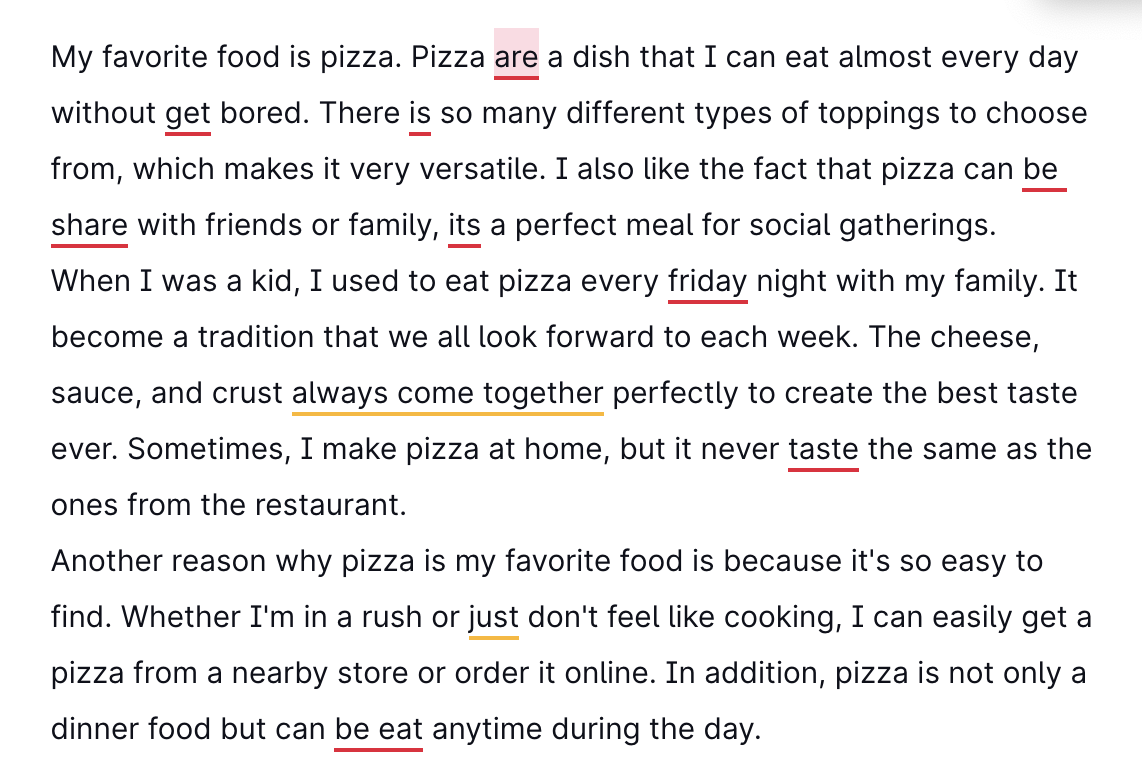}
\caption{Example essay generated with instructions to simply include mistakes}
\label{figSampM}
\end{figure}

One of the simplest ways to include errors in an Agent's writing is to specify them directly in the prompt.
The following is an possible example prompt.
\begin{lstlisting}
Write an essay with grammatical errors answering the following question:
What is your favorite food?
\end{lstlisting}
An example of an such essay is shown in Fig \ref{figSampM}.

As demonstrated above, it is possible to include grammatical errors in the essay.
However, the mistakes generated by this method are not based on the actual user's writing, and it cannot be guaranteed that the generated essay matches the specific user's proficiency level.
In addition, providing a sample of the user's writing and asking the AI an essay at the same proficiency level remains challenging. A sample prompt is shown below, the user's essay is shown in Fig. \ref{figUserE}, and the generated result is shown in Fig. \ref{figCompE}.
\begin{lstlisting}
You are pretending to be a different fictional character who has similar English skills to the person who wrote the given essay. Use the same level of English proficiency, but write a new and original essay that answers the question as if you were this fictional person. Make sure the content of your essay is completely different from the given essay, but keep the language skills and writing style consistent with the given text. Your essay should be written in 6 to 10 sentences and around 70 to 130 words.
\end{lstlisting}

\begin{figure}[tbp]
  \begin{minipage}{0.48\linewidth}
    \centering
    \includegraphics[width=\linewidth]{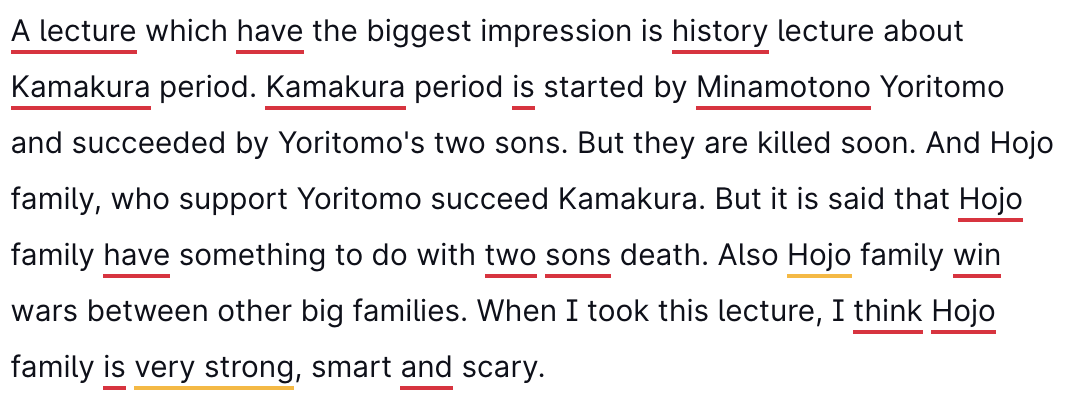}
    \caption{Sample user essay with grammatical errors}
    \label{figUserE}
  \end{minipage}
  \hfill
  \begin{minipage}{0.48\linewidth}
    \centering
    \includegraphics[width=\linewidth]{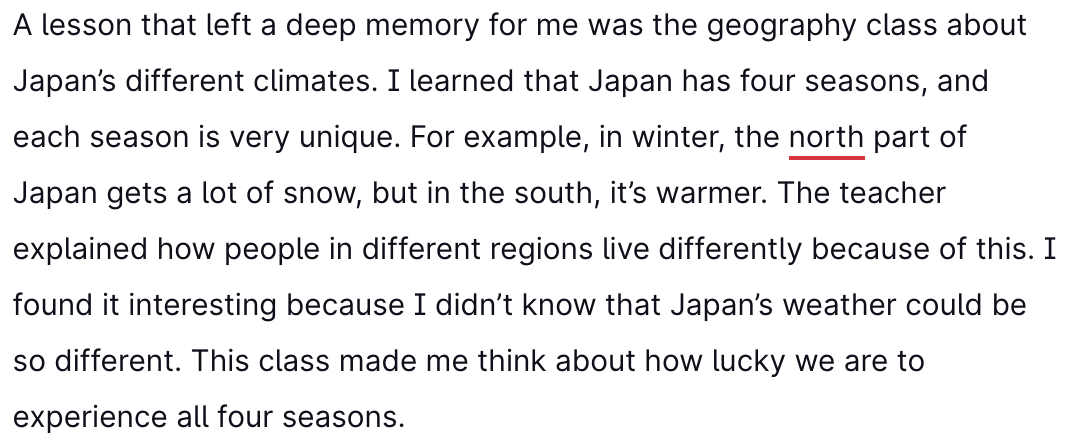}
    \caption{Sample essay generated by comparison method}
    \label{figCompE}
  \end{minipage}
\end{figure}

Therefore, it is challenging for an AI agent to generate sentences that closely match the user's proficiency level. In this study, we use the previously shown prompt, which instructs the agent to mimic the user’s proficiency level, as the basis for the comparison method. We propose a new method that improves this by enabling more accurate replication of the user's proficiency.

\section{Method}
In this section, we present a method that accurately reflects errors in the agent's essays based on the user's proficiency level.
We propose a method inspired by Prompt Insertion\cite{kaneko2023prompt} to develop a mechanism for the LCAA that can replicate mistakes commonly made by learners in English composition. Prompt Insertion has been shown to be effective in explaining grammatical errors in a task called Grammatical Error Correction (GEC), with the aim of applying it to English education. Hereafter in this paper, grammatical errors will be referred to as “Grammatical Errors” or “Errors” to distinguish them from "Mistakes", which refer to general user-made errors during learning.

\begin{figure}[tbp]
\centering
\includegraphics[width=0.8\linewidth]{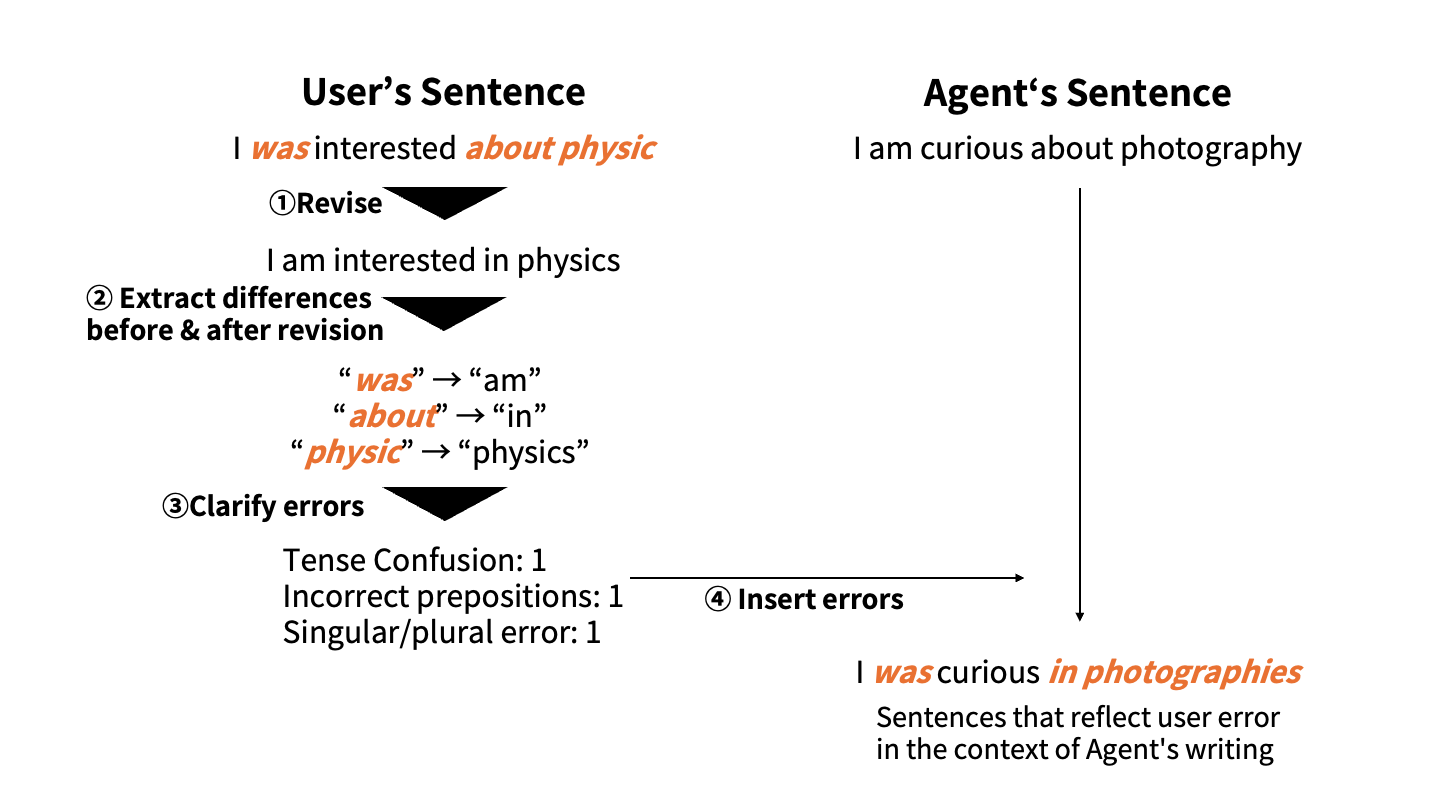}
\caption{Illustration of the proposed method using a simple example}
\label{fig1}
\end{figure}

According to \cite{kaneko2023prompt}, Prompt Insertion generates grammatical explanations in multiple stages and can be considered a form of Chain of Thought\cite{wei2022chain}. Details are provided below.
Fig. \ref{fig1} shows the correction process using the proposed method.

\subsection{Processing Steps}
This study proposes the following four-step flow. Example prompts are also provided below.
\subsubsection{Correct the user's English text}

\begin{lstlisting}
Correct any grammatical or structural errors in the sentences.
\end{lstlisting}

In this step, the AI corrects incorrect parts (i.e., Errors) of the user's input sentences. The process of identifying which parts of the input sentence are incorrect is equivalent to evaluating the sentence.
As described in \cite{oda2017pigai}, existing automatic evaluation systems base their calculations on similarity to native speaker sentences using a large corpus of data. However, in this study, we assume that LLM has already learned the characteristics of native speaker sentences, allowing and let the AI agent evaluate the input sentences.

\subsubsection{Create a list of changes before and after correction}

\begin{lstlisting}[mathescape=true]
List the changes made in the corrected essay compared to the original, in the format 'original text' $\rightarrow$ 'corrected text'. Each item in the list should be extracted only for the changed portion and not the entire sentence. In other words, there should be no multiple changes within a single list item. Only list changes where there is a meaningful grammatical or structural correction, and ignore formatting differences.
\end{lstlisting}

This step lists corrections as 'original text' $\rightarrow$ 'corrected text,' focusing only on meaningful grammatical changes. 
The purpose of this procedure is to make it easier to analyze what mistakes were made in the original user’s text by listing the differences before and after each correction, for each corrected part.

\subsubsection{Clarify the types of errors and count them}

\begin{lstlisting}
Explain the mistakes made in the corrected essay. The list of changes is a list of areas that need to be corrected for the text. Categorize each correction by specific grammatical element (e.g., tense, word choice, subject-verb agreement) and count how many errors were made. For example, the list should look like:
- Subject-verb agreement error: 1
- Incorrect article usage: 2
At the end of the list, provide a total count, e.g., "Total corrections: 8".
Be as specific and precise as possible in your explanation of the errors.
\end{lstlisting}

This procedure is to count up the number of errors for each kinds in the list made in the previous step.

\subsubsection{Insert errors in AI agents' essay}

\begin{lstlisting}
In the given essay, mix the specified number of errors of the specified type into the essay to produce a grammatically incorrect essay. Do not include any explanations, headers, or lists of errors in your output. Only output the grammatically incorrect essay itself, and ensure the number of errors matches the total specified.
\end{lstlisting}

This step instructs the AI agent to write the same number of mistakes of that type as were counted by the AI agent in the previous step.

\section{Experiment}
\subsection{Experimental conditions}
Experiments were conducted to compare the effectiveness of the proposed method with the comparative method. The proposed method, as described in Section IV, extracts errors from the user’s essay and reflects them in the agent’s generated essay. In contrast, the comparison method, explained in Section III, involves providing the agent with the user’s essay and having it write a new essay that mimics the user’s proficiency. We applied both methods to various essays from participants and compared their effectiveness.

In this study, eight Japanese participants, either currently enrolled in or graduated from a four-year college or university, wrote essays on four simple topics, providing a total of 32 essays. These essays were analyzed using Grammarly for (a) basic grammatical errors (Errors), and (b) essay scores (Quality), with the latter being a built-in metric provided by Grammarly to assess the overall quality of the writing.
The number of sentences assigned to both the participants and the AI agents was matched, ensuring that the number of errors corresponded directly to the error percentage within the entire essay.
We analyzed the essays generated by both methods: the proposed method, which extracts and reflects user errors through a multi-step process (described in Section IV), and the comparison method, where the agent writes a new essay based on the user's proficiency. The same analysis was applied to these essays as was done with the human participants' data. In this study, we used GPT-4-o for the model and LangChain for agent management. The essay topics were designed to be simple and familiar to encourage natural writing and to focus on grammatical and stylistic accuracy. The topics used are listed below.
\begin{enumerate}
\item 
Please describe a country or region you would like to visit.
\item 
Please introduce your favorite movie.
\item 
Explain whether you prefer outdoor activities or indoor play.
\item 
What is the most impressive class you have taken so far?
\end{enumerate}

\subsection{Results}
This section describes the results obtained in this experiment. Fig. \ref{fig2} and Fig. \ref{fig3} below show the average number of errors (Errors) and the average quality rating (Quality) detected by Grammarly in the essays produced by each method. Quality is measured on a 100-point scale.
\begin{figure}[tbp]
\centering
\includegraphics[width=0.48\linewidth]{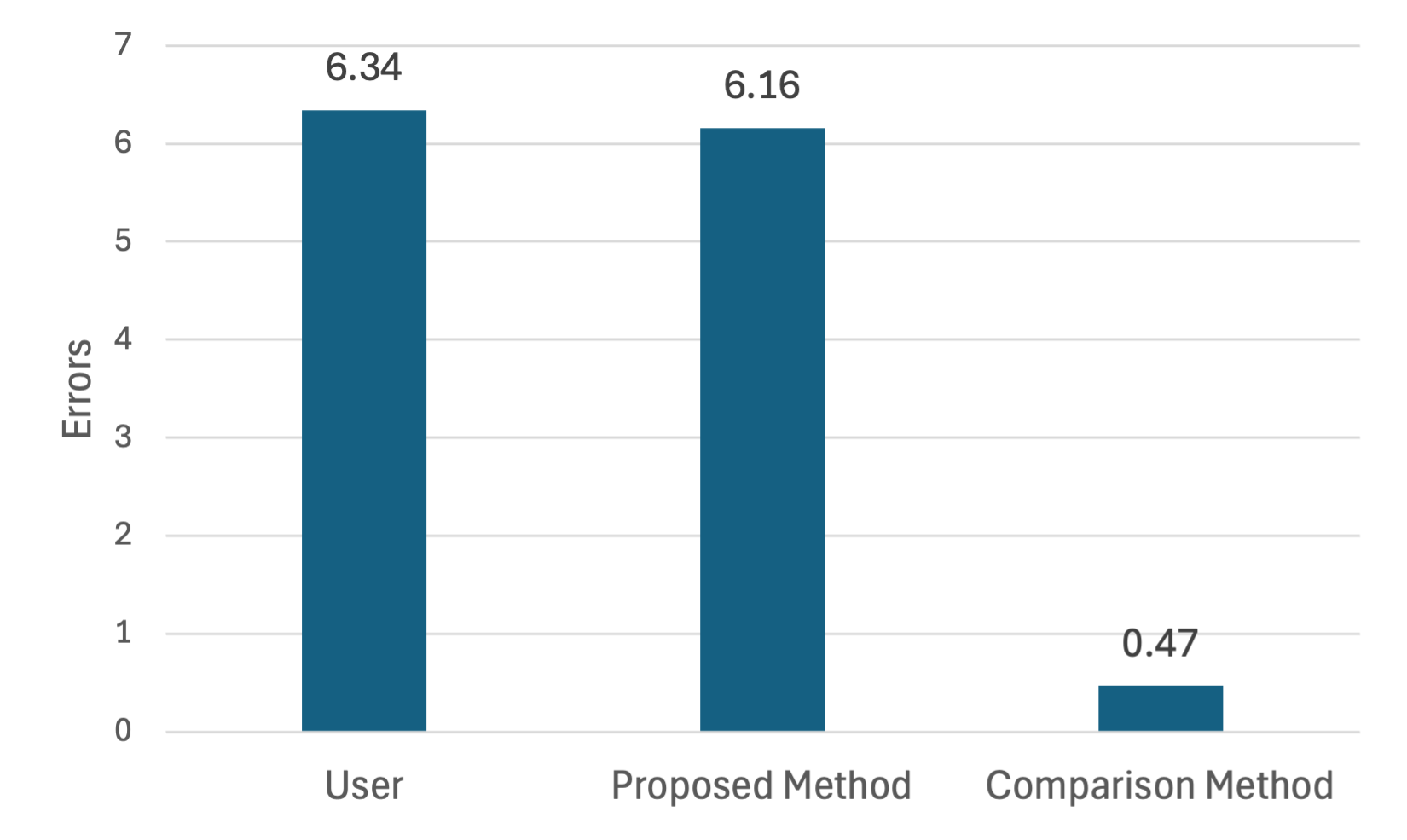}
\caption{Average of the number of errors in each type of sentence}
\label{fig2}
\end{figure}

The average number of errors in user sentences was approximately 6.34. The proposed method resulted in an average of 6.16 errors per sentence, whereas the comparison method averaged only 0.47 errors. Therefore, the number of errors in sentences produced by the proposed method more closely matches the error count in user sentences compared to the comparison method.

\begin{figure}[tbp]
\centering
\includegraphics[width=0.48\linewidth]{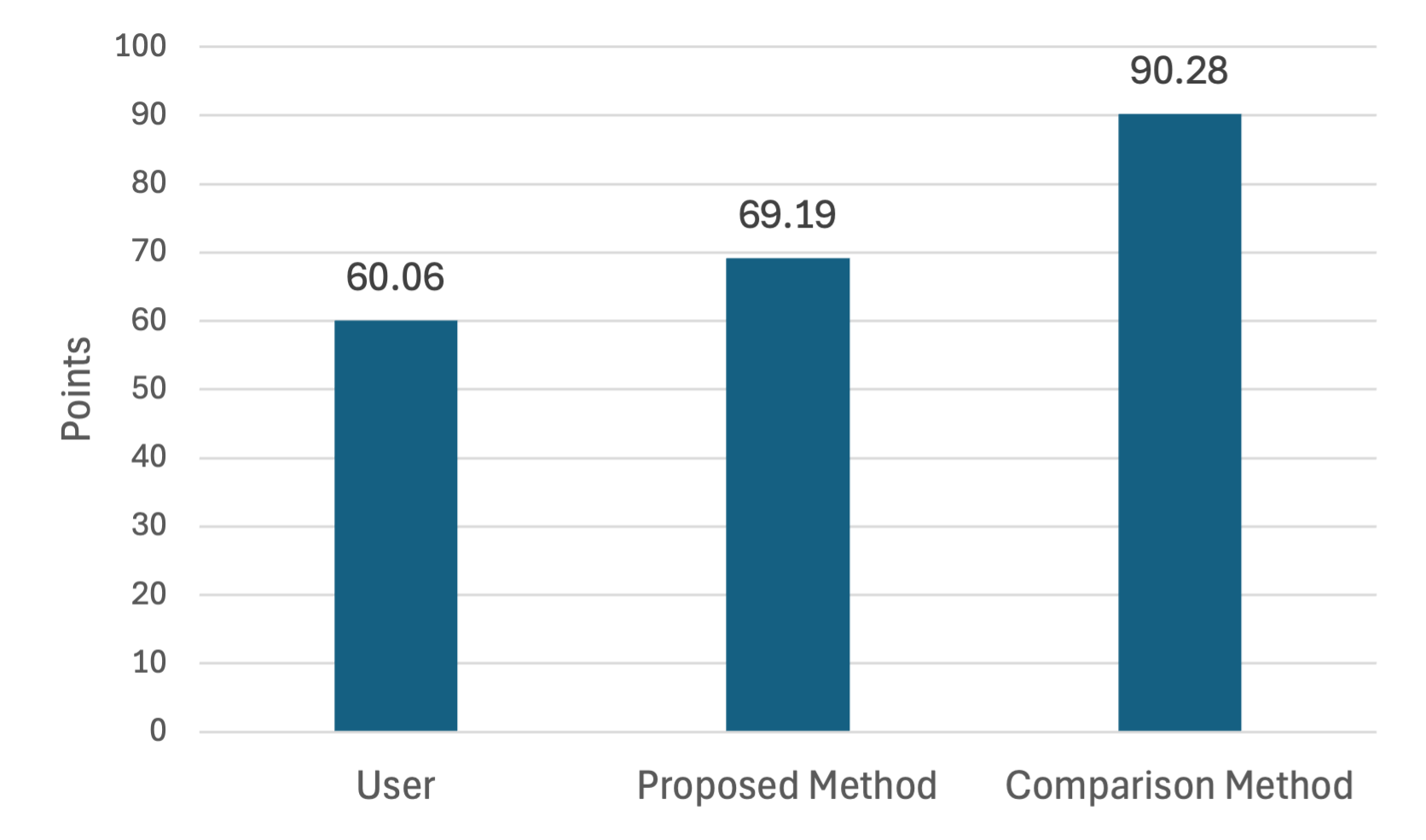}
\caption{Average of the quality score in each type of sentence}
\label{fig3}
\end{figure}

In terms of sentence quality, the average score of the user's sentences is 60.06, while the proposed method averages is 69.19 and the comparison method averages score is 90.28. Based on these averages, the sentence quality produced by the proposed method is closer to that of the user's sentences than that of the comparison method.

\begin{figure}[tbp]
\centering
\includegraphics[width=0.48\linewidth]{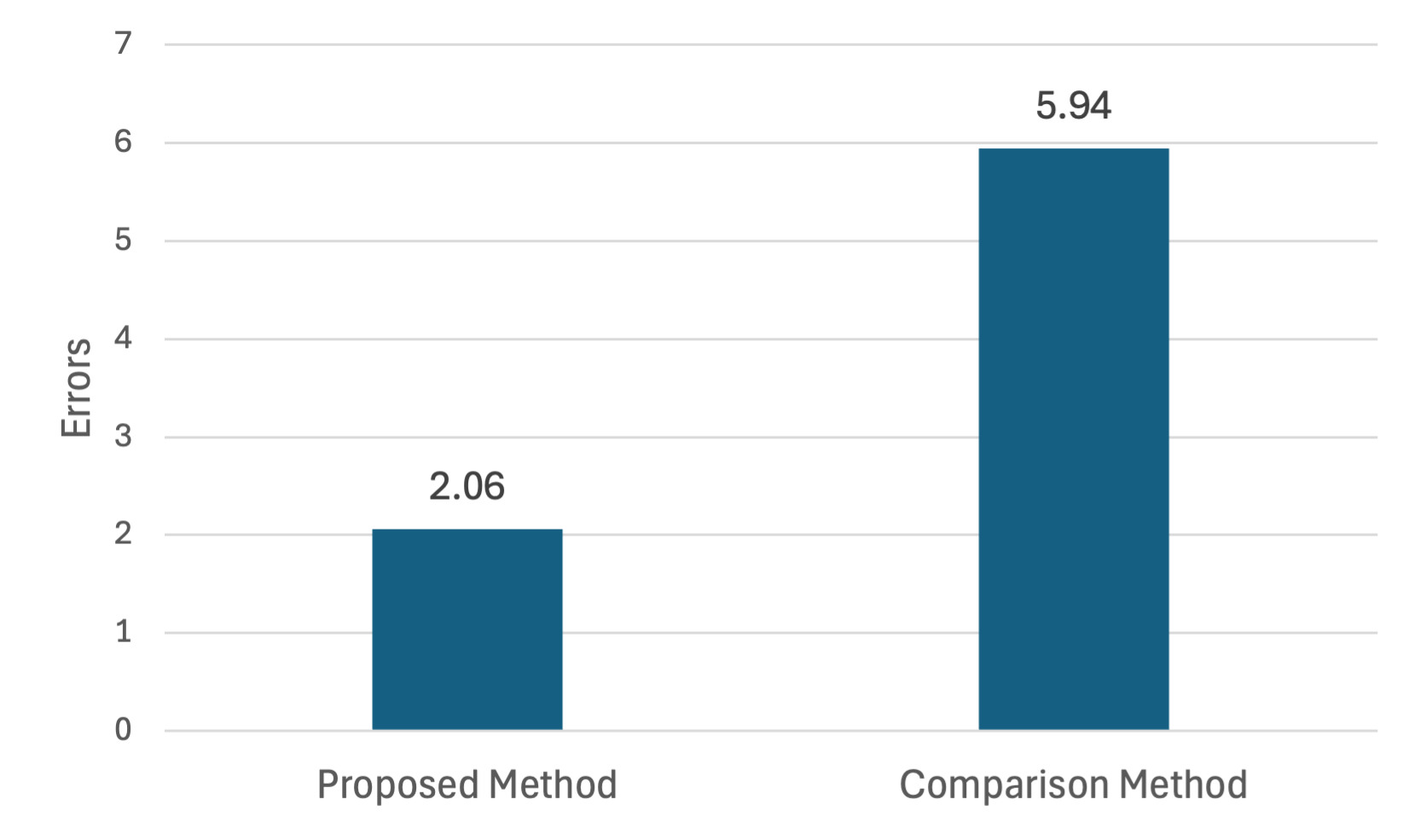}
\caption{Average of the absolute difference in the number of errors}
\label{fig4}
\end{figure}

Fig. \ref{fig4} shows the average absolute error difference between the user's sentence and both the proposed and comparison methods. In this study, the proposed method resulted in an average error difference of 2.06, compared to 5.94 for the comparison method. 
Thus, the proposed method reduces the error difference to less than half that of the comparison method.

\begin{figure}[tbp]
\centering
\includegraphics[width=0.48\linewidth]{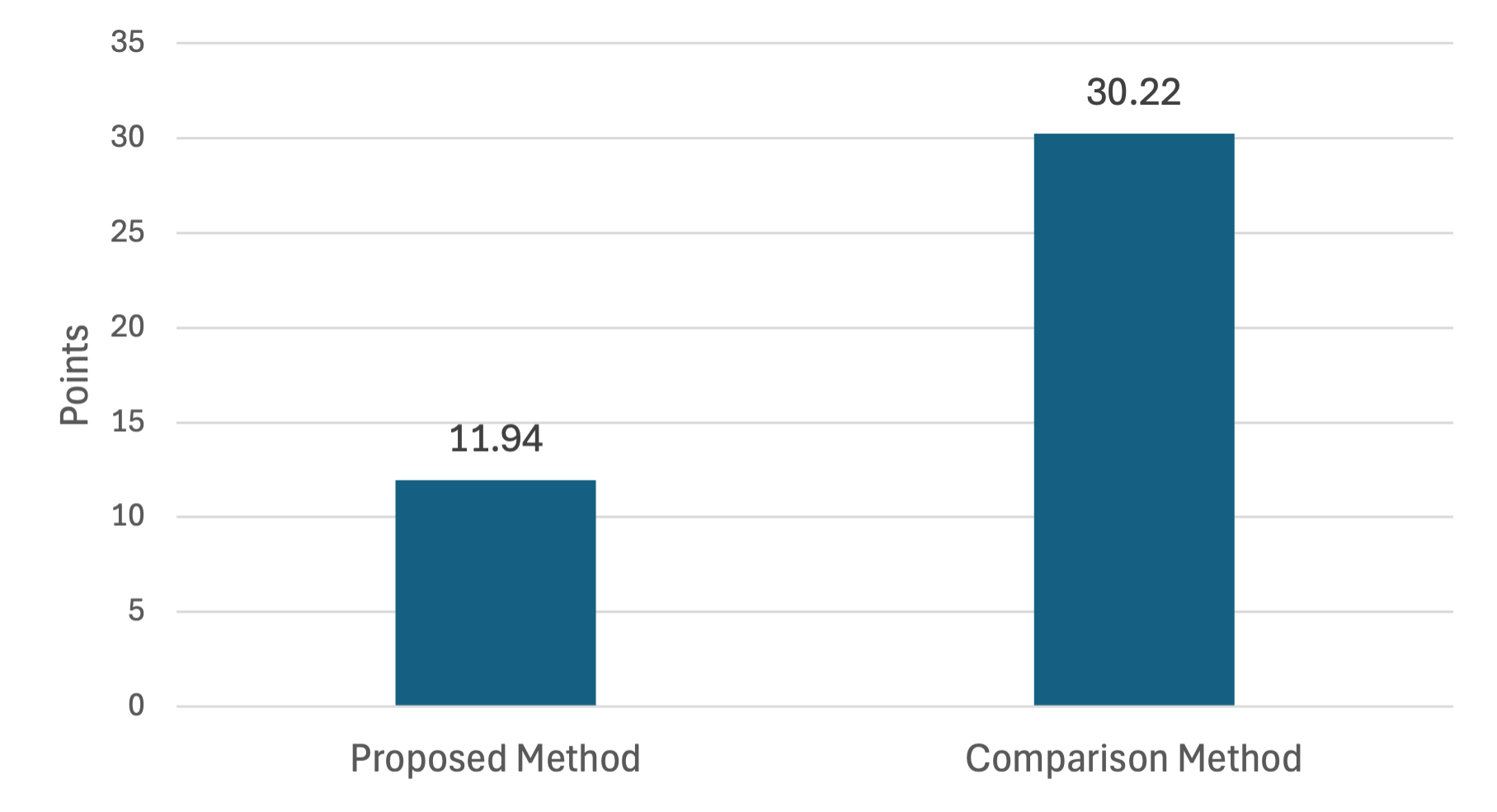}
\caption{Mean of absolute differences in quality indicators}
\label{fig5}
\end{figure}

Fig. \ref{fig5} similarly shows the mean of the absolute differences in Grammarly's measures of writing quality.
The average absolute difference between the measures of the quality of the user sentences and the proposed method was on average approximately 11.94, while the absolute difference between the quality of the user sentences and the comparison method was 30.22.

In summary, in terms of the absolute difference of each index and the user's statement, the experimental method shows about half the difference compared to the comparison method in the number of errors (Errors) and the quality index (Quality).

Tab. \ref{tab1} below shows the results of t-tests on the raw values and absolute differences for both the experimental and comparative methods.
The result indicate that the proposed method significantly outperforms the comparison method in both the number of errors (Errors) and the Grammarly Score (Quality), a qualitative measure of errors and quality, under the conditions of this experiment.

\begin{table}[tbp]
\caption{Results of t-test (P-value) of Difference from User's Score}
\begin{center}
\begin{tabular}{|c|c|}
\hline
\textbf{Tested Items} & \textbf{P-value} \\
\hline
\textbf{Errors} & $3.002 \times 10^{-8}$ \\
\hline
\textbf{Quality} & $1.731 \times 10^{-10}$ \\
\hline
\end{tabular}
\label{tab1}
\end{center}
\end{table}

Based on t-test results, we calculated Cohen's d to measure the effect size between the proposed method and the comparison method in terms of both the number of errors and the sentence quality. The results are shown in Tab. \ref{tab2} below. According to Cohen's d, an effect size of 0.2 is considered small, 0.5 medium, and 0.8 large. 

\begin{table}[tbp]
\caption{Cohen's d Values for Error Count and Sentence Quality Between Methods}
\begin{center}
\begin{tabular}{|c|c|c|}
\hline
\textbf{Tested Items} & \textbf{Cohen's d} \\
\hline
Errors & 3.70 \\
\hline
Quality & 2.91 \\
\hline
\end{tabular}
\label{tab2}
\end{center}
\end{table}

In this study, the calculated Cohen's d values for the number of errors (3.70) and sentence quality (2.91) indicate a very large effect size. Therefore, based on Cohen's d, it can be concluded that the proposed method demonstrates a significantly greater effect compared to the comparison method.

Based on the results of the statistical analysis above, the sentences generated by the agent using the proposed method are presented below. The user's text contains grammatical errors, such as part-of-speech errors, tense errors, singular/plural errors, and word choice errors, which are reflected in the sentences generated by the agent.

\begin{framed}
\noindent
\textbf{Example sentence created by Agent with proposed method:} \\
One place I have always wanted to visits is New Zealand. The stunning landscapes, from lush forests to dramatic mountains, seems like they belong in a dream. I am particular drawn to the South Island, where I could explore the fjords and hike through scenic trails. The culture also fascinates me, specially the indigenous Māori traditions. I imagine the people to be warm and welcoming, adding to the overall charms. Exploring the wildlife, like seeing penguins or dolphins, would makes the trip unforgettable.
\end{framed}

The above results show that our proposed method can generate sentences that are clearly closer to the human learner's sentences in terms of both the number and quality of errors than the conventional method for using AI agents, which simply provides the user's sentences and learner attributes, and prompts the AI to imitate the user's writing.

\section{Conclusion}
In this paper, we proposed an AI-based peer learning environment that employs a Learning Companion AI Agent (LCAA) as a peer in English composition. The LCAA generates errors similar to those made by the user, enabling more effective peer learning through user-like mistake generation. To evaluate this approach, we conducted experiments comparing the proposed method with a comparison method that instructed the AI to mimic the user’s proficiency level when generating errors. The results showed that the proposed method better reflected the user’s error patterns and aligned more closely with their proficiency than the comparison method. While the proposed method effectively mimicked user-like mistakes, we tested the AI agent’s ability to handle sentences with grammatical errors detectable by Grammarly, which would typically lead to ungrammatical sentences. However, more complex issues, such as word choice or idiomatic usage, required further analysis. Additionally, since this study was conducted on the number of errors in the sentences, a detailed breakdown of the types of errors needs further verification. Moreover, as this experiment only addressed grammatical errors in English composition, it remains untested how the method would perform when imitating sentence-level errors caused by factual inaccuracies, syntactic errors, or variations in syntax or expression.

It is noteworthy that Grammarly's qualitative indicators closely matched human sentences via the proposed method, even though the overall sentence level was not specifically mimicked by variations in expression.

\section*{Acknowledgment}
This work was supported by JST BOOST, Japan Grant Number JPMJBS2406.
We would like to thank the eight people who participated in the experiment, as well as seminar colleagues who gave us various advice on our research.

\bibliographystyle{IEEEtran}  
\bibliography{references}  


\end{document}